\documentclass[letterpaper, 10 pt, conference]{ieeeconf-modified}  

\IEEEoverridecommandlockouts                              

\usepackage[numbers]{natbib}
\usepackage{graphicx}
\graphicspath{ {./figures} {../plots} {./plots}}
\usepackage{siunitx}
\usepackage{algorithm}
\usepackage{algorithmic}
\usepackage{amsmath} 
\usepackage{amssymb}  
\usepackage{textcomp}
\usepackage{hyperref}
\usepackage[capitalize]{cleveref}
\usepackage{makecell}
\usepackage{capt-of}
\usepackage{booktabs}

\DeclareMathOperator*{\argmin}{arg\,min}

\crefname{equation}{}{}
\newcommand{\gitlink}{https://aidx-lab.org/manipulation/iros24}

\title{\LARGE \bf
Learning a Shape-Conditioned Agent for \\Purely Tactile In-Hand Manipulation of Various Objects
}
\author{Johannes Pitz$^{1*}$ \;\; Lennart Röstel$^{1*}$ \;\; Leon Sievers$^1$ \;\; Darius Burschka$^2$ \;\; Berthold Bäuml$^1$
}

\begin{document}

\twocolumn[{%
        \renewcommand\twocolumn[1][]{#1}%
        \maketitle
        \begin{center}
            \centering
            \vskip -0.3cm
            \includegraphics[width=\textwidth]{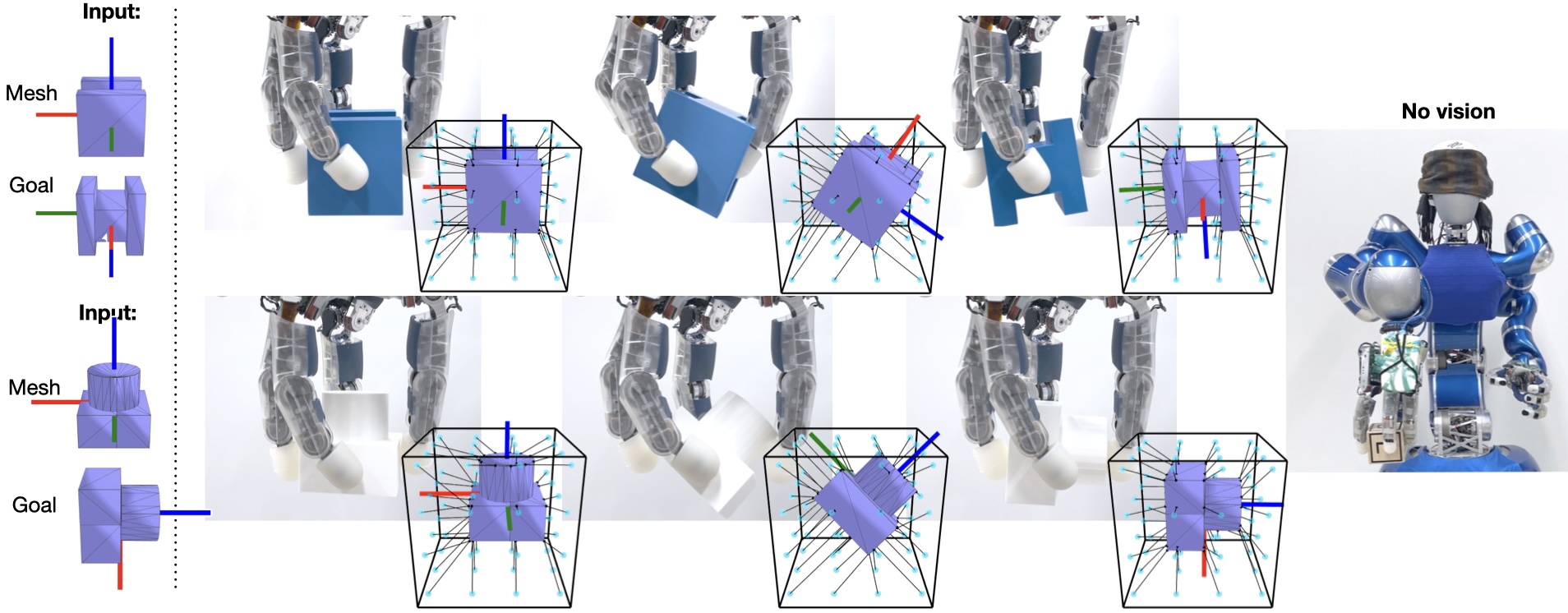}
            \captionof{figure}{Purely tactile, shape-conditioned in-hand reorientation with the torque-controlled DLR-Hand II~\cite{Butterfass2001} and Agile Justin~\cite{Bauml2014-cr} (right).
            Conditioned on a mesh input in the initial pose and a goal orientation (left), our learned agent autonomously reorients various objects towards the target without visual information or supporting surfaces (top: training object 7, bottom: out-of-distribution object 11). 
            Shape information is encoded as vectors to the mesh, transformed by an estimate of the current object pose, which is predicted by a learned state estimator from the history of force/position measurements. }
            \label{fig:title_figure}
        \end{center}%
    }]

{\let\thefootnote\relax\footnote[0]{
 $^*$ Equal contribution 
 \newline $^1$ Learning AI for Dextrous Robots Lab (\href{https://aidx-lab.org/}{\scriptsize\texttt{aidx-lab.org}}), Technical University of Munich, Germany, and DLR Institute of Robotics \& Mechatronics (German Aerospace Center)
 \newline $^2$ Machine Vision and Perception Group, Technical University of Munich
 \newline Contact:{\tt\scriptsize{ \{johannes.pitz|lennart.roestel\}@tum.de}}
 }
}

\thispagestyle{empty}
\pagestyle{empty}

\vspace{-1.5mm}
\begin{abstract}
    Reorienting diverse objects with a multi-fingered hand is a challenging task.
    Current methods in robotic in-hand manipulation are either object-specific or require permanent supervision of the object state from visual sensors.
    This is far from human capabilities and from what is needed in real-world applications.
    In this work, we address this gap by training shape-conditioned agents to reorient diverse objects in hand, relying purely on tactile feedback (via torque and position measurements of the fingers' joints).
    To achieve this, we propose a learning framework that exploits shape information in a reinforcement learning policy and a learned state estimator. 
    We find that representing 3D shapes by vectors from a fixed set of basis points to the shape's surface, transformed by its predicted 3D pose, is especially helpful for learning dexterous in-hand manipulation. 
    In simulation and real-world experiments, we show the reorientation of many objects with high success rates, on par with state-of-the-art results obtained with specialized single-object agents.
    Moreover, we show generalization to novel objects, achieving success rates of $\sim$90\% even for non-convex shapes.  \\
    Website: \href{\gitlink}{\scriptsize\texttt{\gitlink}}
\end{abstract}

\section{Introduction}
In-hand reorientation is a key skill for industrial applications and many downstream robotic tasks. 
Yet, robustly manipulating objects with arbitrary shapes in a controlled manner remains an open problem, especially without supporting surfaces~\cite{Sievers2022}.
Performing this task purely tactile, i.e., without a vision sensor that provides the measurements of the absolute object pose, is even more challenging and requires deliberate manipulation strategies that avoid uncontrolled object movements~\cite{Rostel2023-nc}.
Given that in industrial settings the 3D mesh of the object to be manipulated is often available, using this shape information explicitly for in-hand manipulation seems promising. 
Yet, in particular, for learning-based approaches, the question arises of which 3D representation best facilitates the learning process and generalization to new objects.
In this paper, we study the effectiveness of different shape representations in the context of reinforcement learning-based in-hand reorientation of various objects without visual sensors.
Using our insights, we extend our previous object-specific training method~\cite{Rostel2023-nc,Pitz2023-ra} to shape-conditioned agents, where a single policy and estimator can deal with various objects, including novel objects unseen during training.

\subsection{Related Work}

Reinforcement learning has emerged as a promising approach for solving complex in-hand manipulation tasks. 
Most prior works either train policies to rotate continuously about a single axis~\cite{Sievers2022,Qi2022-va,Khandate23}, or for multiple axes but for a specific object~\cite{OpenAI2018-yi, openai2019rubiks,Handa2022-rc,Pitz2023-ra,Rostel2023-nc,Khandate24}.
It was recently shown that it is possible to train a single policy for manipulating multiple different objects, including objects unseen during training, by utilizing visual information~\cite{Chen2023-mq,Qi2023-th,Yuan23}, where \citet{Chen2023-mq} even consider a goal-oriented task.

By providing the policy with depth images as input, information about the object shape is implicitly available (up to occlusions).
The depth input modality usually prompted the use of point clouds for representing shapes, where \citet{Yuan23} additionally show that touch information can be integrated into the point cloud.
However, the additional need for a camera sensor can be a significant drawback in many applications.

Since we are not using vision sensors, we are not compelled to use point clouds. 
Instead, we propose to use a shape representation inspired by basis point sets (BPS)~\cite{Prokudin2019-sg}, that computes the shape-conditioned observation based on the estimated object pose.

\citet{chen2021system, Chen2023-mq} showed that, in simulation, it is possible to train goal-oriented policies with diverse objects even without any shape information when provided the ground truth object pose.
Contrary to this, our findings demonstrate that in a simulation setup conducive to robust sim2real transfer of purely tactile agents, incorporating additional shape information proves advantageous, even when the agents are trained with the ground truth object pose.
Moreover, we show that shape information is even more beneficial in the challenging setting where the absolute object pose is not given but has to be estimated from tactile feedback.

In the purely tactile setting, \citet{Yin2023-tw} previously showed in-hand rotation of different objects about fixed axes using touch sensors. 
However, the work doesn't address the challenge of autonomously reaching desired goals.
Instead, it relies on a human operator to judge the current object orientation and decide on the next rotation axis.
In contrast, in our setting of goal-conditioned reorientation, the object pose needs to be estimated from the history of tactile feedback, requiring more deliberate manipulation strategies that avoid losing track of absolute object orientation~\cite{Rostel2023-nc}.

\subsection{Contributions}

\begin{itemize}
    \item We show for the first time purely tactile goal-oriented in-hand manipulation of diverse objects with a single, shape-conditioned agent, i.e., without external sensors.
    \item We provide a comparison of different shape representations for learning dexterous in-hand reorientation and find a favorable shape-encoding: vectors from a fixed set of points (BPS \cite{Prokudin2019-sg}) to the object mesh transformed according to its 3D pose.
    \item We show that the learned shape-conditioned agents
    \begin{itemize}
        \item manipulate many objects with similar success rates as agents that are trained for the given object only.
        \item generalize to novel objects outside the training distribution.
        \item demonstrate successful zero-shot sim2real transfer matching the performance in simulation.
    \end{itemize}
\end{itemize}

\pagebreak

\section{Method}
\label{sec:method}

\subsection{Problem Formulation}
We study the problem of reorienting objects from a grasped configuration to a given target orientation $R_g\in\mathrm{SO}(3)$ in-hand, without supporting surfaces (also no palm support).
Our proposed method assumes torque and position feedback of a torque-controlled hand as the only measurements, with no cameras or other external sensors.
In this work, we additionally require the 3D mesh of the object in its canonical pose $\mathcal{M}$ (i.e., at $x = 0$ and $R = I$), and its initial orientation $R_0$ relative to the hand frame. 
Subsequent measurements of object poses are not required.

\subsection{Representing Shapes for In-Hand Manipulation}
We consider an agent conditioned on a \textit{shape encoding}
\begin{align}
    \mathcal{S} = \mathcal{S}(x, R, \mathcal{M}) ,
\end{align}
which is a representation of the object shape as a function of the canonical mesh $\mathcal{M}$ and the object position $x\in \mathbb{R}^3$ and orientation $R\in \mathrm{SO}(3)$.
The pose could, for example, be obtained from a visual tracking system or, as in our experiments, a learned tactile estimator (cf. \cref{sec:estimator}).
We would like the shape representation to:
\begin{itemize}
    \item contain information about the object surface geometry that is necessary to perform in-hand manipulation.
    \item facilitate learning of a reinforcement learning agent.
    \item allow trained agents to generalize to new object shapes.
\end{itemize}
While we also consider other shape representations for comparison later in this work (\cref{sec:experiments_oracle}),
in the following, we focus on Basis Point Sets (BPS)~\cite{Prokudin2019-sg}, an effective encoding method for learning on 3D shapes.
\subsubsection{Basis Point Sets (BPS)}
BPS define a fixed set of $N_b$ basis points $\{p_k\in \mathbb{R}^3\}_{k = 1,..., N_b}$, for each of which the vector $v_k\in \mathbb{R}^3$ to the closest point on the mesh surface $\text{surf}(\mathcal{M})$ is computed:
\begin{equation}
    \begin{aligned}
        &v_k = p^*_k - p_k \\
        p^*_k =& \argmin_{p \in \text{surf}(\mathcal{M})} ||p - p_k|| .
    \end{aligned}
\end{equation}
The BPS encoding vector $V$ is then simply the concatenation of these vectors (or its associated magnitudes):
\begin{align}
    V = \left[v_1, ... , v_{N_b}\right] \in \mathbb{R}^{3 N_b} .
\end{align}
Unlike point cloud data, the BPS encoding has intrinsic order and thus does not require permutation-invariant network architectures, such as PointNets~\cite{Qi2016-qr}.

In the robotic context, BPS have been used for representing obstacles in neural motion planning~\cite{Tenhumberg2022-li} 
and for representing object shapes for grasp prediction~\cite{Winkelbauer2022-uh}. 

\subsubsection{Basis Point Sets for dynamic objects}
In this work, we use the BPS encoding to represent the object shape for the policy, as well as for the state estimator.
To this end, we extend the notion of BPS to the dynamic case, where the basis point encoding is now additionally a function of the time-varying object pose $(x, R)$:
\begin{align}    
	v_k = b(k, x, R, \mathcal{M}) = \big(\argmin_{p \in \text{surf}(\mathcal{M}, x, R)} ||p - p_k||\big) - p_k ,
\end{align}
where the closest point is now calculated relative to the surface of $\mathcal{M}$ translated by $x$ and rotated by $R$.
If $p_k$ lies inside of the transformed mesh, we set $v_k = 0$.
We also define
\begin{equation}
    \begin{aligned}
        \label{eq:bps_vec}
        V_t &= \mathcal{B}(x_t, R_t, \mathcal{M}) \\
        &= \Bigl[b(1, x_t, R_t, \mathcal{M}), ..., b({N_b}, x_t, R_t, \mathcal{M})\Bigr] .
    \end{aligned}
\end{equation}
Note that each basis point vector $v_k$ falls into exactly one of two categories: It either points to an edge of the mesh or is aligned with a surface normal. 
In both cases, we expect $v_k$ to contain valuable information, especially for contact-rich in-hand manipulation tasks.

In our experiments, we place $N_b=4^3$ basis points uniformly on a grid [-7\,cm, 7\,cm]$^3$.

\begin{figure}[t!]
    \centering
    \vspace{3mm}
    \includegraphics[width=0.48\textwidth]{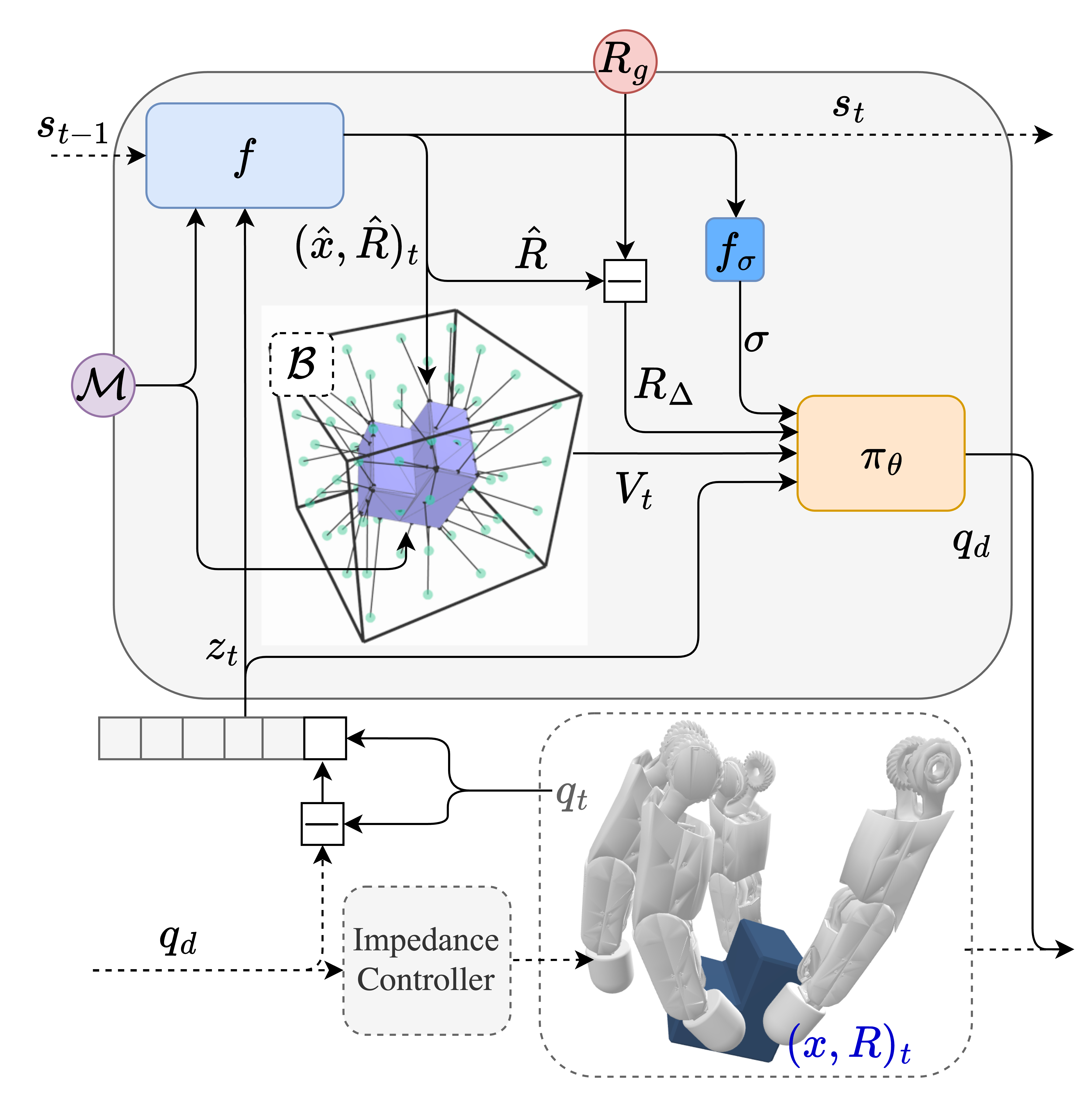}
    \caption{Shape-conditioned agent control architecture. The tactile state estimator $f$ predicts the system state $s_t$ recursively (see \cref{fig:estim_cell}).
    Based on the estimated object pose $(\hat{x},\hat{R})$ and the given mesh $\mathcal{M}$, the shape respresentation $\mathcal{S}_t=\mathcal{B}(\hat{x}_t, \hat{R}_t, \mathcal{M})=V_t$ is computed in each timestep $t$.
    This is fed to the policy $\pi$ together with the relative rotation to the goal $R_{\Delta}$, a stack of joint measurements $z_t$, and the predicted uncertainty $\sigma$, to produce actions $q_d$.
    These actions, produced at a frequency of $10$Hz, are then low-pass filtered and given to an impedance controller at $1000$Hz for controlling the hand.}
    \label{fig:policy_cell}
\end{figure}
\subsection{Shape-Conditioned Agent}

Our agent consists of a policy $\pi$ and a state estimator $f$ (cf. \cref{sec:estimator}).
The policy outputs $N_\text{dof} = 12$ target joint angles
\begin{align}
    q_d \sim \pi(R_{\Delta}, \mathcal{S}, z, \sigma) \in \mathbb{R}^{N_\text{dof}} .
\end{align}
As input the policy receives the relative rotation to the goal $R_{\Delta}$, a representation of the object shape $\mathcal{S} = \mathcal{S}(x, R, \mathcal{M})$ and a measurement vector $z$. 
For our agent, $z$ consists of a stack of $N_\text{s}=6$ joint measurements $q$ and control errors $e=q_d-q$ sampled at \SI{60}{Hz} for a window of $0.1$ seconds:
\begin{align}
z_t = [q_{t-(N_\text{s}-1)}, e_{t-(N_\text{s}-1)},... , q_{t}, e_{t}] \in \mathbb{R}^{2N_\text{dof} \cdot N_\text{s}} .
\end{align}
For low-level control, we use a high-fidelity impedance controller. This allows the agent to infer contacts from the control error even without direct access to the torque signal.
Additionally, if a state estimator is used to compute $R_{\Delta}$ and $\mathcal{S}$, we include their predicted uncertainties $\sigma$ (see \cref{sec:estimator}) in the input to the policy.
An overview of the proposed agent is shown in~\cref{fig:policy_cell}.

\subsection{Learning the State Estimator}
\label{sec:estimator}
\begin{figure}[t]
    \centering
    \vspace{3mm}
    \includegraphics[width=0.4\textwidth]{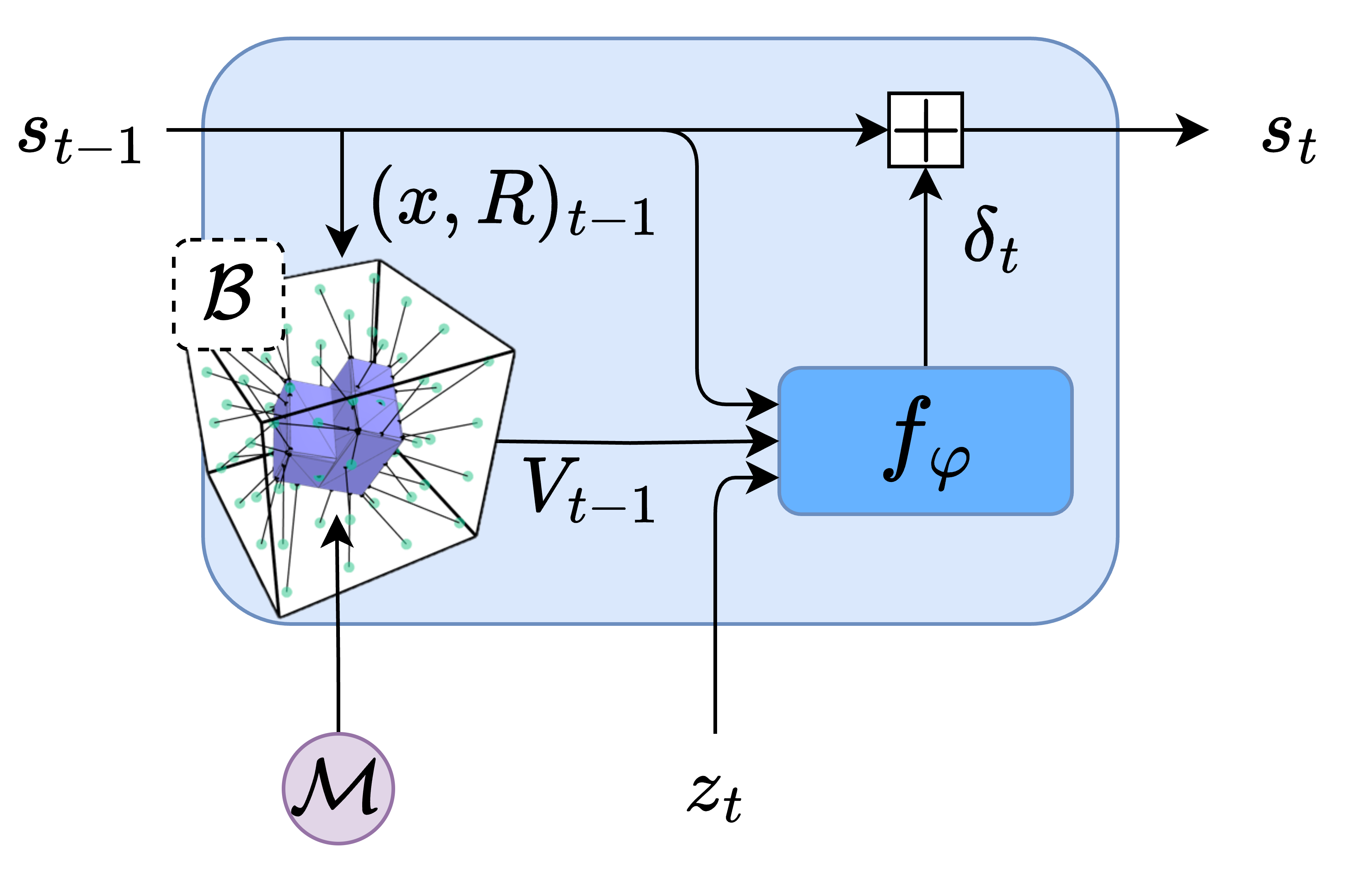}
    \caption{Estimator cell $f$. In each timestep, the BPS feature vector $V_{t-1}$ is computed from the estimated object pose $(\hat{x}_{t-1}, \hat{R}_{t-1})$ and mesh $\mathcal{M}$.
    The learned function $f_{\varphi}$ predicts the residual in object state $\delta_t$ from $V_{t-1}$, the previous latent state $l_{t-1}$, and the joint measurements $z_t$.
    Finally, state $s_t$ is obtained by (generalized) addition of $\delta_t$ to $s_{t-1}$.}
    \label{fig:estim_cell}
\end{figure}
In the line of prior works~\cite{Rostel2022-gu,Pitz2023-ra,Rostel2023-nc}, we train a state estimator $f$ that predicts the object state based purely on the history of joint measurements, provided through the torque and position sensors.
We extend this approach by incorporating shape information within the estimator's recurrent architecture. 

Similarly to \cite{Rostel2023-nc}, the recurrent state is composed of estimated object position $\hat{x}$, estimated orientation $\hat{R}$, and a latent state vector $l$:
\begin{align}
    s_t = (\hat{x}_t, \hat{R}_t, l_t) \in \mathbb{R}^3 \times \mathrm{SO}(3) \times \mathbb{R}^n .
\end{align}
$\hat{R}_0$ is initialized with a known rotation matrix $R_0$, while $\hat{x}_0$ and $l_0$ are learned.
Based on $\hat{x}_t$ and $\hat{R}_t$, in each timestep the BPS feature vector is computed as $V_t = \mathcal{B}(\hat{x}_t, \hat{R}_t, \mathcal{M})$.
Together with $l_t$ and the joint measurements $z_t$, it is then passed to a learnable function $f_{\varphi}$ that predicts the residual in object state $\delta_t \in \mathbb{R}^{3 + 3 + n}$ for the next timestep.
An overview of the state estimator cell is shown in~\cref{fig:estim_cell}.

For optimizing the learnable parameters of $f_\varphi$, we treat the basis point set calculation as part of the computation graph, enabling us to backpropagate through $\mathcal{B}$ recurrently.
For supervised training, we propose a loss function based on the differences between the ground truth basis point vectors $v_k$ and the predicted ones:
\begin{align}
    \label{eq:bps_dist}
    d_v(\hat{x}, \hat{R}, \mathcal{M}) = \frac{1}{N_b}\sum_{k=1}^{N_b} \left| ||v_k|| - ||b(p_k, \hat{x}, \hat{R}, \mathcal{M})|| \right| .
\end{align}
Note that in~\cref{eq:bps_dist}, we penalize the difference in the magnitudes of the basis point vectors rather than the directional components, as the distance is continuous w.r.t. changes in object pose.
Because the above distance metric is invariant to object symmetries, we additionally include a term that penalizes the angle between the absolute predicted and ground truth object rotation $d(\hat{R}, R)$. 
The overall loss function $\mathcal{L}$ for the state estimator is the negative log-likelihood of the distances under a Normal distribution with sigmas predicted by a feed-forward network $f_{\sigma}$ from the latent state $l_t$:
\begin{equation}
    \begin{aligned}
        \sigma_{v_t}, \sigma_{R_t} &= f_{\sigma}(l_t) \\ 
        \mathcal{L} = -\frac{1}{T}\sum_{t=0}^{T} &\log \varphi\left(d_v(\hat{x}_t, \hat{R}_t, \mathcal{M})| \sigma_{v_t}\right) \\
        &+ \log \varphi\left(d( \hat{R}_t, R_t)| \sigma_{R_t}\right) .
    \end{aligned}
\end{equation}
Here, $\varphi(\cdot|\sigma)$ is the pdf of a Normal distribution with zero mean and standard deviation $\sigma$.

\subsection{Learning a Policy / Reinforcement Learning Task}

Our reinforcement learning task is similar to the one described in~\cite{Rostel2023-nc}, where we give a positive reward when the angle to the goal~$\theta_{t}=d(R_t, R_{g})$ is smaller than the previous~$\theta_{t-1}$. 
We also penalize deviations of the object position~$x_t$ and joint angles $q$ from their initial values~$x_0$ and $\bar{q}_0$, respectively.
In summary, the reward function is
\begin{align}
  r_t = \quad\; &\lambda_{\theta} \min \left( \theta_{t-1} - \theta_{t}, \theta_\text{clip} \right) \nonumber \\
             - &\lambda_{x} ( \| x_t - x_0\| - \| x_{t-1} - x_0\| ) \\
             - &\lambda_{q} \| ( q_t - \bar{q}_0 )^4 \|_1 \nonumber .
\end{align}
The coefficients are $\lambda_{\theta}= 1$, $\lambda_{x}= 8$, $\lambda_{q} = \frac{1}{6}$, $\theta_\text{clip} = 0.1$\,rad. 
Clipping the rotation reward forces the policy to rotate the object slowly and, therefore, more controlled, which is preferable for a robust sim2real transfer.

For each episode, we uniformly sample a goal orientation $R_{g} \sim \mathrm{SO}(3)$.
We generate at least one valid initial state for each simulation environment by spawning a randomly orientated object in the hand and closing the fingers around it using a grasping heuristic(see also~\cite{Rostel2023-nc}).
The reorientation is considered a success if $\theta_T=d(R_T, R_{g}) < 0.4$ after $T=10$ seconds.

\subsection{Training the Agent}

We train the policy and the estimator using Estimator-Coupled Reinforcement Learning (ECRL)~\cite{Rostel2023-nc}.
The policy is trained by reinforcement learning, specifically Proximal Policy Optimization (PPO)~\cite{Schulman2017proximal}, concurrently with the state estimator, which is trained to regress the ground truth object state from the simulator. 
Critically, during training, the policy receives the predicted state from the estimator, which leads the policy to avoid actions that would lead to unpredictable object movements~\cite{Rostel2023-nc}.
This is in contrast to Student-Teacher approaches~\cite{Qi2023-th,Chen2023-mq}, which may produce actions only appropriate when provided with absolute state information (as from vision-based systems).

We provide detailed hyperparameter choices for the policy, the estimator, and the ECRL training process on the project website.

\subsection{Implementation}

We employ a GPU-accelerated simulation environment based on Isaac Sim \cite{IsaacSim}, where we model the hand and objects with substantial domain randomization (object mass, friction, control parameters, measurement noise) to enable sim2real transfer (see~\cite{Sievers2022,Rostel2023-nc}). 
Our PPO implementation is based on \citet{rl-games2021} and the policy network uses the D2RL \cite{dense_d2rl} architecture.
For training on multiple objects, we equally distribute the object set among 4096 simulation instances, which are run in parallel on a single NVIDIA T4 GPU.
The BPS calculations are performed by a custom Warp Kernel~\cite{warp2022}, which enables parallelization across basis points, object meshes, and timesteps. 
Furthermore, the kernel is differentiable with respect to its inputs, enabling the estimator to backpropagate through the BPS calculations. 

\begin{figure*}
    \centering
    \vspace{3mm}
    \includegraphics{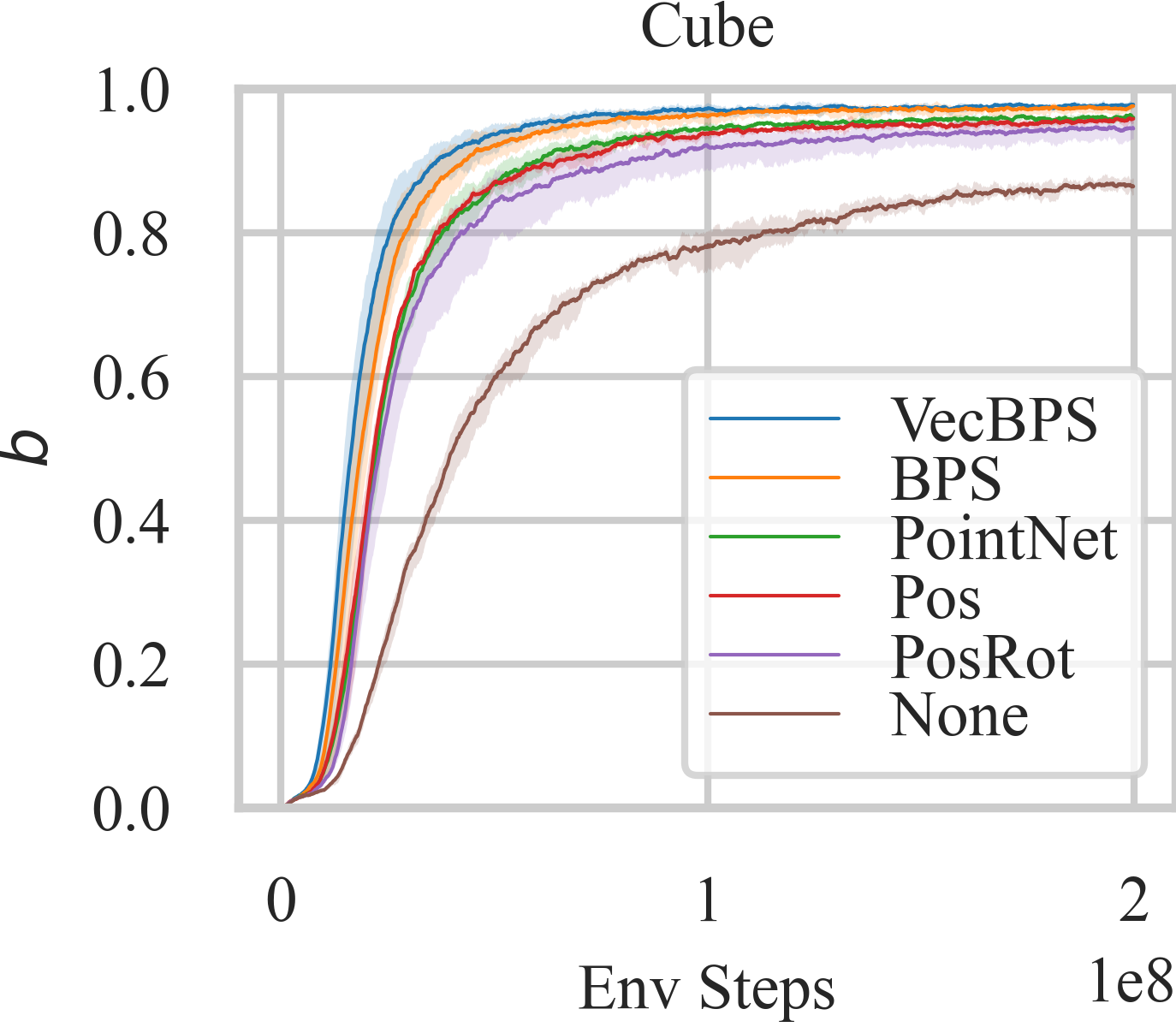}
    \includegraphics{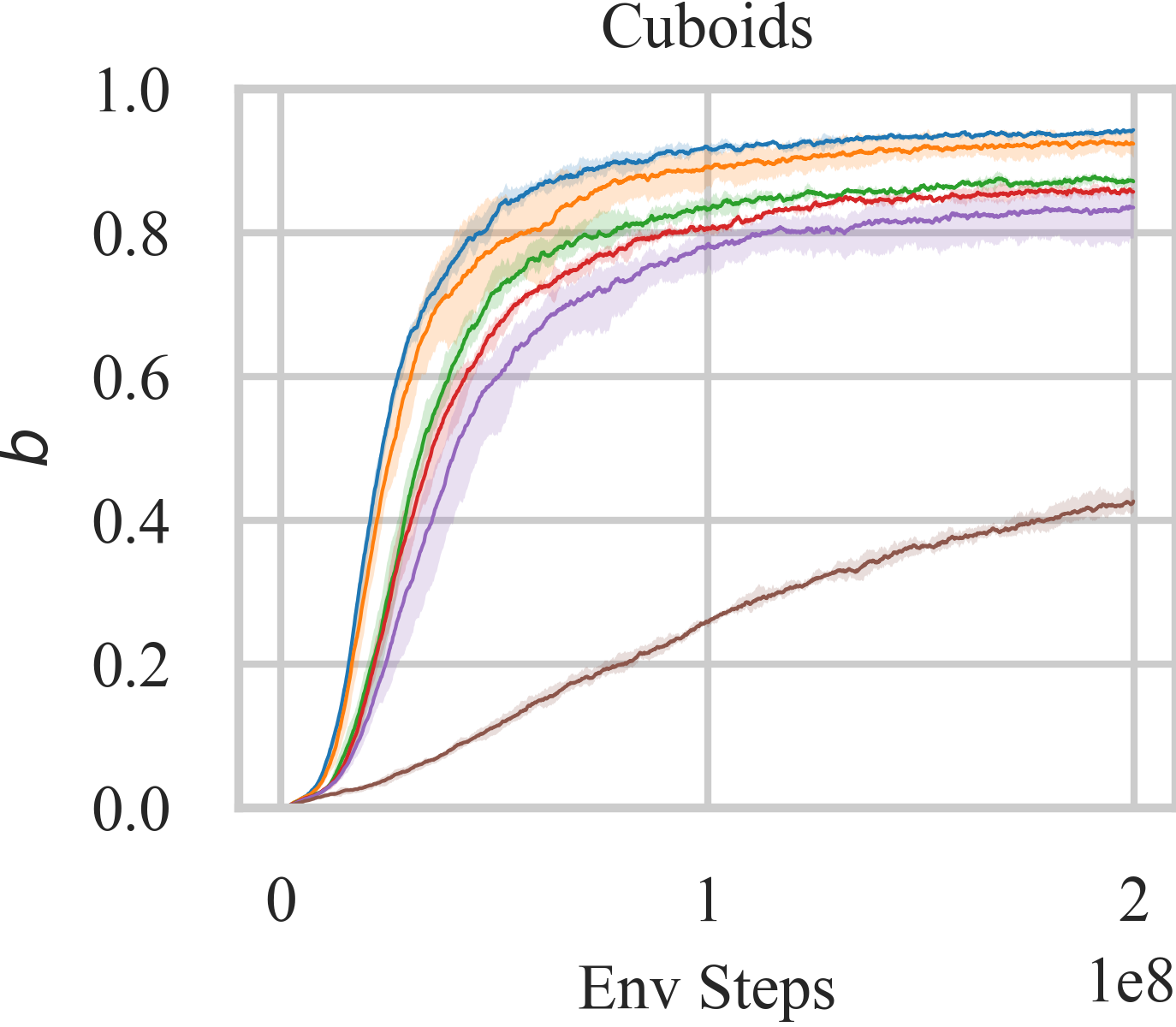}
    \includegraphics{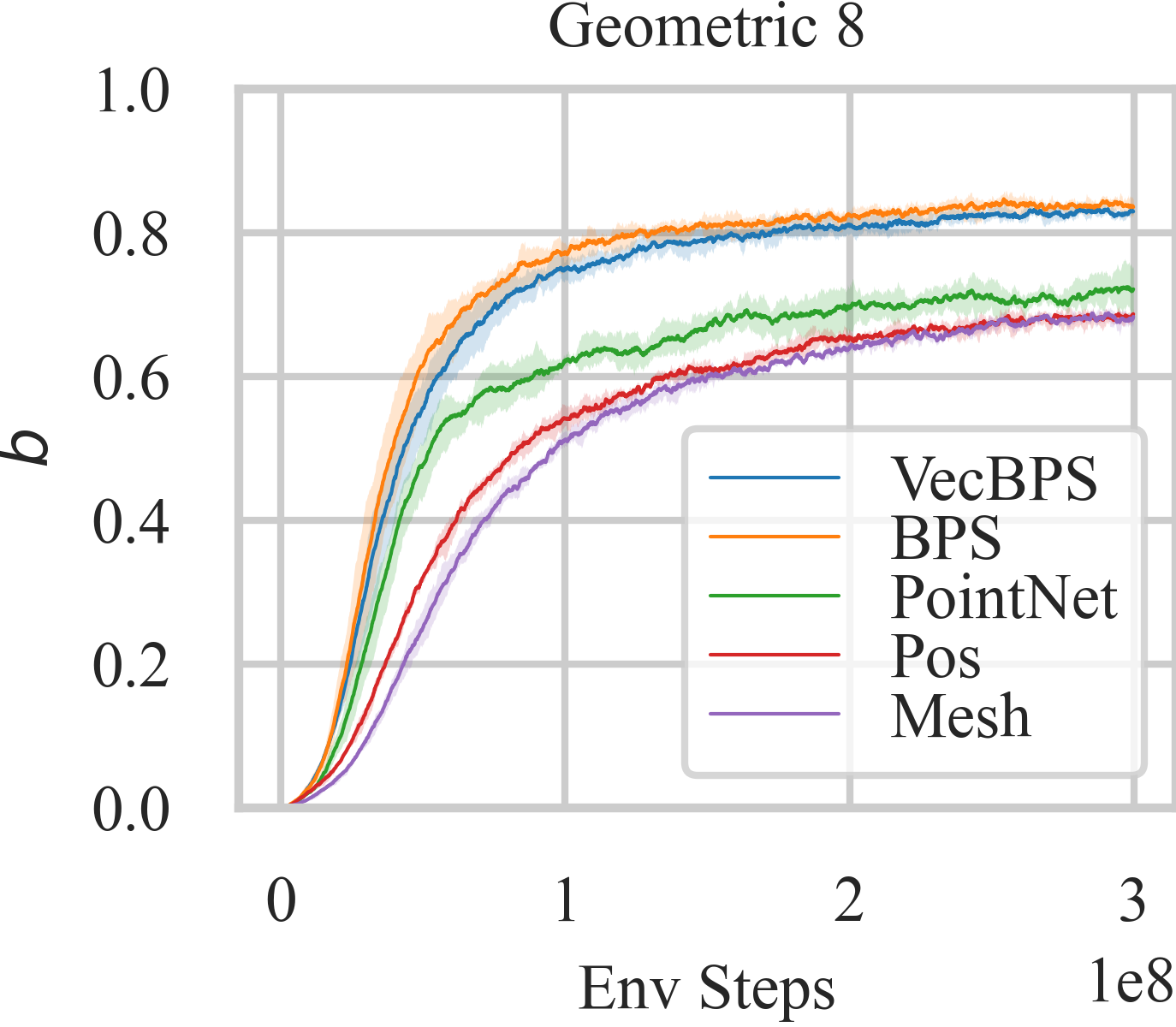}
    \caption{
        Success rates $b$ during training of \textit{oracle} agents with different shape representations, plotted against the total number of environmental steps.
        \textbf{Left:} Training on a single object (Cube, index 8). \textbf{Middle:} Training on cuboids with randomized axis-independent scaling in [4.5\,cm, 9\,cm]. \textbf{Right:} Training on the \textsc{Geometric 8} object set with randomized scaling.
        Each line is the mean over three training runs, with shaded areas covering the min and max.
        We smooth the (binary) success signal for the individual runs.
    }
    \label{fig:learning}
\end{figure*}

Overall, training takes around 60 hours for the tactile agents (policy and estimator).
The \textit{oracle} agents introduced in \cref{sec:experiments_oracle} step faster and converge earlier.
The runs reported in \cref{fig:learning} train for 7-11 hours.

\section{Experiments \& Analysis}
\label{sec:experiments}
In this section, we report experiments in simulation comparing different methods for conditioning the reinforcement learning agent on shape information and real-world experiments proving the sim2real transfer of our simulation setup.

\subsection{Object Datasets}

\begin{figure}[h]
    \centering
    \includegraphics[width=.48\textwidth]{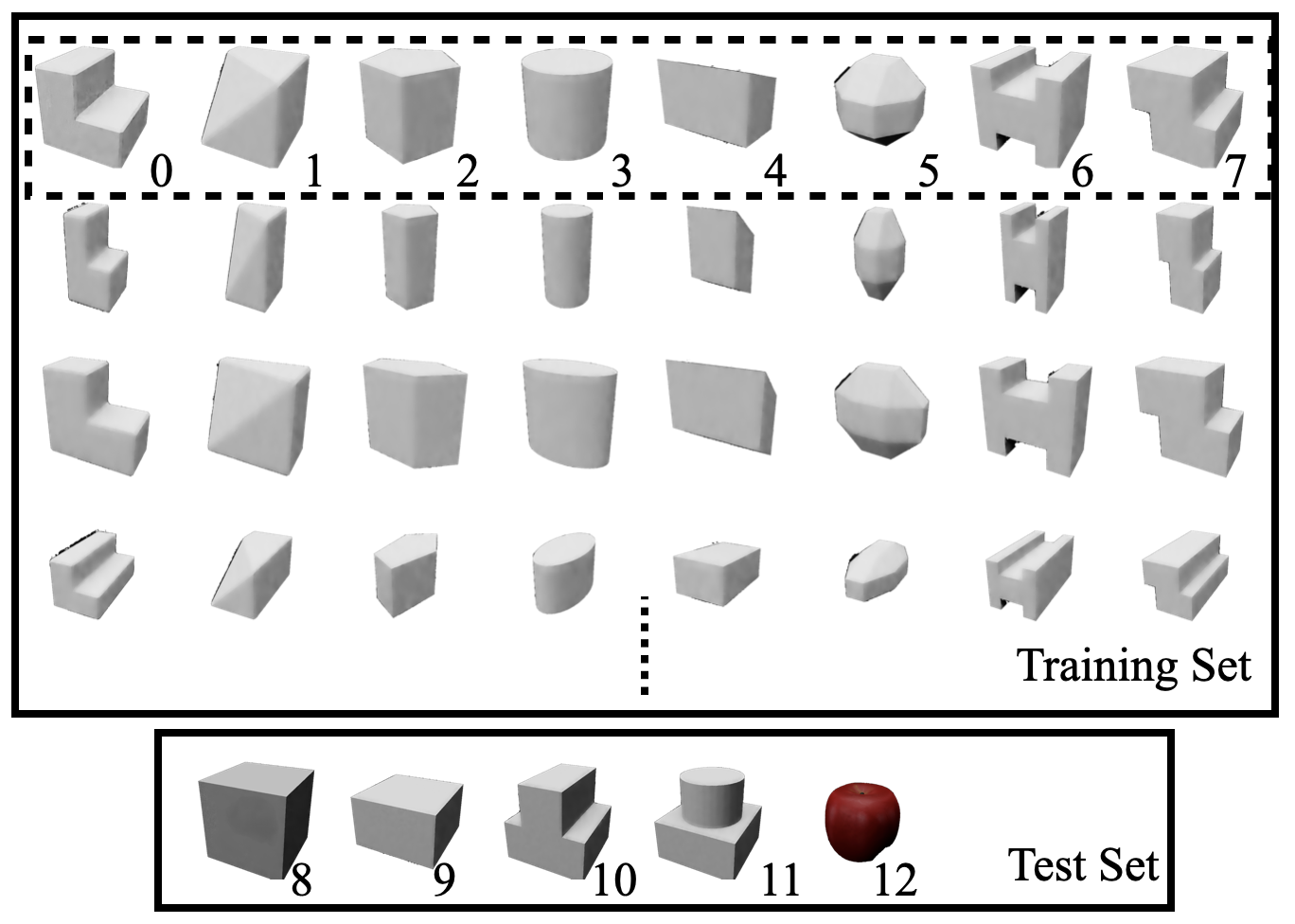}
    \caption{
        Object sets used in the experiments: objects 0-7 (\textsc{Geometric 8}) are designed to have different geometric properties (convex, non-convex, varying number of edges, angle between adjacent surfaces, etc.).
        The training set consists of these eight geometric objects with randomized axis-independent scaling in [4.5\,cm, 9\,cm], effectively creating a much larger set of training objects.
        Out-of-distribution (\textsc{OOD}) test objects 8-12 are not present in the training set and are used for assessing generalization capabilities.
        Object 12 is the YCB apple \cite{YCB}. 
    }
    \label{fig:object_set}
\end{figure}

We train and evaluate the shape-conditioned agents on different sets of objects, shown in~\cref{fig:object_set}.

\subsection{Comparing Shape Representations (Oracle)}
\label{sec:experiments_oracle}

To assess the importance of shape information for learning in-hand reorientation by reinforcement learning,
we first consider a setting where the object pose is known to the policy, and no tactile estimator is needed.
In this setting, the \textit{oracle} agent (opposed to the tactile agent) is synonymous with the policy.
We compare the performance of policies with the following object encodings $\mathcal{S}$:
\begin{itemize}
    \item \textit{VecBPS} - the BPS encoding $\mathcal{S} = V$
    \item \textit{BPS} - the magnitudes of the BPS vectors 
    \item \textit{PointNet} - point cloud processed using a PointNet-encoder~\cite{Qi2016-qr}, but keeping the orientation dependence by dropping the input transform as \citet{Yuan23}.
    \item \textit{PosRot} - the object pose $\mathcal{S} = (x, R)$
    \item \textit{Mesh} - the object pose and the mesh in canonical pose $\mathcal{S} = (x, R, \mathcal{M})$
    \item \textit{Pos} - only the object position $\mathcal{S} = x$
    \item \textit{None} - no object information 
\end{itemize}

To make the comparison as fair as possible within our compute limitations we reuse the same hyperparameters as in our previous work \cite{Rostel2023-nc}, which were tuned for the \textit{Pos} variant and only tune additional hyperparameters such as the resolution of the BPS and the number of points (256) and the encoder network (linear layer with 64 units) of the PointNet.

In \cref{fig:learning}, we show the learning curves for the different inputs when training on object datasets of increasing complexity.
The BPS-based encodings generally outperform the other input representations for all object sets.
While the performance benefit of the BPS encoding is marginal when training on a single object (cube, 8), it becomes more pronounced with increasing complexity of the object set.
Moreover, in our experiments, the BPS-based encodings consistently outperform the PointNet encoding, which provides only moderate improvements over omitting shape information entirely.

We found it interesting to see that providing the policy with the orientation of the object (\textit{PosRot}) seems to increase the variance across runs but does not improve the performance over only providing the position (\textit{Pos}).
However, that should not be surprising because the orientation is meaningless without defining the object's origin and its general shape.
For example, for two cuboids with extents (5, 5, 8)\,cm and (5, 8, 5)\,cm respectively, the same orientation will result in different configurations, which will generally only "confuse" the policy.
Although this could, in principle, be mitigated by additionally providing the canonical mesh alongside the pose (\textit{Mesh} representation), we find that this does not improve performance when testing the hypothesis on the \textsc{Geometric~8} object set.
This underlines the effectiveness of transforming the BPS representation according to the current pose.

The position proved helpful even though the same problem of defining the origin exists.
But unlike orientation changes, which can be inferred from the goal orientation delta $R_\Delta$, changes in position provide additional information.
Additionally, all objects are centered around the origin.
Therefore, the absolute position indicates when the object might be close to dropping or sliding up too high.

\subsection{Benchmarking Purely Tactile Shape-Conditioned Agents}
\label{sec:experiments_estim}

\begin{figure*}
    \centering
    \vspace{3mm}
    \includegraphics[width=7in]{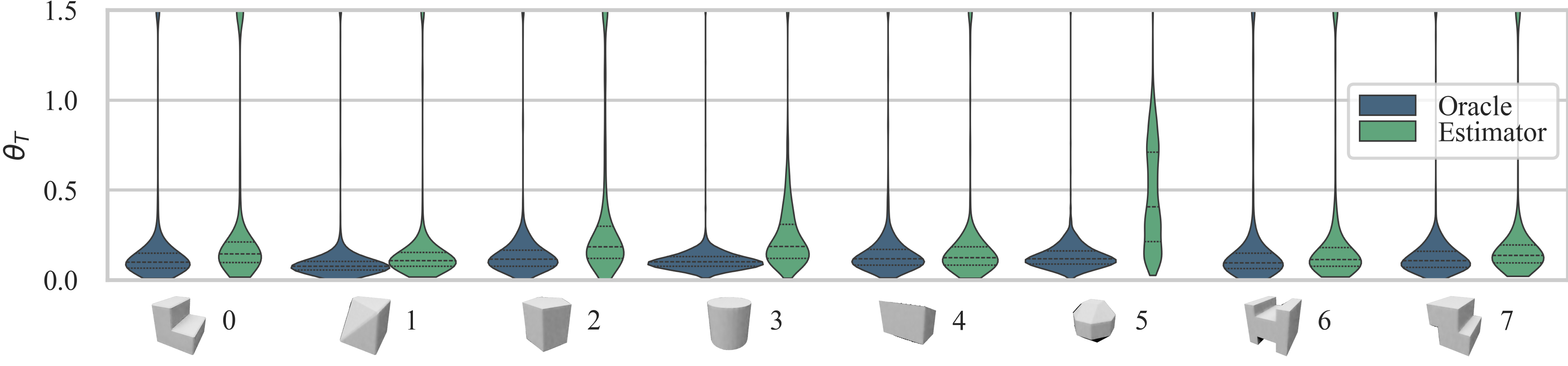}
    \caption{
        Distributions of the final angle $\theta_T$ to the goal orientation for individual objects (no axes scaling). 
        Each violin entails 24 goals for 64 environments (=1536 samples).
        Final angles greater than 1.5\,rad, and episodes where the object was dropped are grouped to the point at the violin's top.
        We compare the \textit{oracle} to the purely tactile agent (estimator), using the \textit{VecBPS} representation for both.
    }
    \label{fig:violinplot}
\end{figure*}

For the remainder of this section, we present results for agents trained with state estimator as described in \cref{sec:method}, where the ground truth object state is not available to the policy but has to be inferred from the history of joint measurements by the estimator.
Unless noted otherwise, we use the \textit{VecBPS} encoding as the shape representation to the policy and estimator and train on the \textsc{Geometric 8} object set with randomized scaling.

We evaluate final performances by measuring the final angle to the goal orientation after 10\,s, and report the success rate $b$ for each object as the fraction of goals reached within a threshold of $0.4$\,rad.
Dropping the objects counts as a failure. 
As in~\citet{Pitz2023-ra,Rostel2023-nc}, we systematically evaluate all 24 goal orientations in a $\pi/2$-discretization of the $\mathrm{SO}(3)$ space (Octahedral group), although our policies are trained to reorient from arbitrary object initializations to any goal orientation in $\mathrm{SO}$(3).

In \cref{fig:violinplot}, we compare the distributions of final angles to the goal orientation for the oracle agent and the tactile agent (with estimator).
While we expected the average angle error to be higher for the tactile agent, the most prominent differences compared to the \textit{oracle} case are observed for objects with less pronounced edges, such as object 3 (the cylinder) or object 5 (the 26 sided "sphere").

\begin{table}
    \centering
    \caption{Success Rates by Object Properties}
    \begin{tabular}{l c c c c c }
        \toprule
        \multicolumn{2}{c}{Objects} & \multicolumn{2}{c}{Oracle} &\multicolumn{2}{c}{Estimator} \\
        Description & IDs & \textit{VecBPS} & \textit{Pos} & \textit{VecBPS} & \textit{Pos}\\
        \midrule
        Convex & 1, 2, 3, 4, 8, 9    & 0.97 & 0.94 & 0.90 & 0.70 \\
        Round & 5, 12                & 0.99 & 0.99 & 0.66 & 0.51 \\
        Non-convex & 0, 6, 7, 10, 11 & 0.95 & 0.88 & 0.90 & 0.68 \\
        \bottomrule
    \end{tabular}
    \label{tab:success_by_property}
\end{table}

In \cref{tab:success_by_property}, we investigate this phenomenon closely by comparing the average success rates of \textit{oracle} and tactile agents (with estimator) with and without shape information for three subsets of our objects.
Note how both \textit{oracle} agents perform exceptionally well on round objects.
The \textit{oracle} agent without shape information scores over $10\%$ higher on the round objects than on the non-convex ones.
However, this trend is entirely reversed for the tactile agents, particularly the one with shape information.
There, the success rate is almost $25\%$ lower on the round objects than on the non-convex ones.
The table shows that shape information is especially useful for manipulating non-convex shapes and confirms that estimating the absolute pose of round objects purely from tactile feedback is very challenging~\cite{Rostel2023-nc}.

\begin{figure}
    \centering
    \includegraphics[width=3.4in]{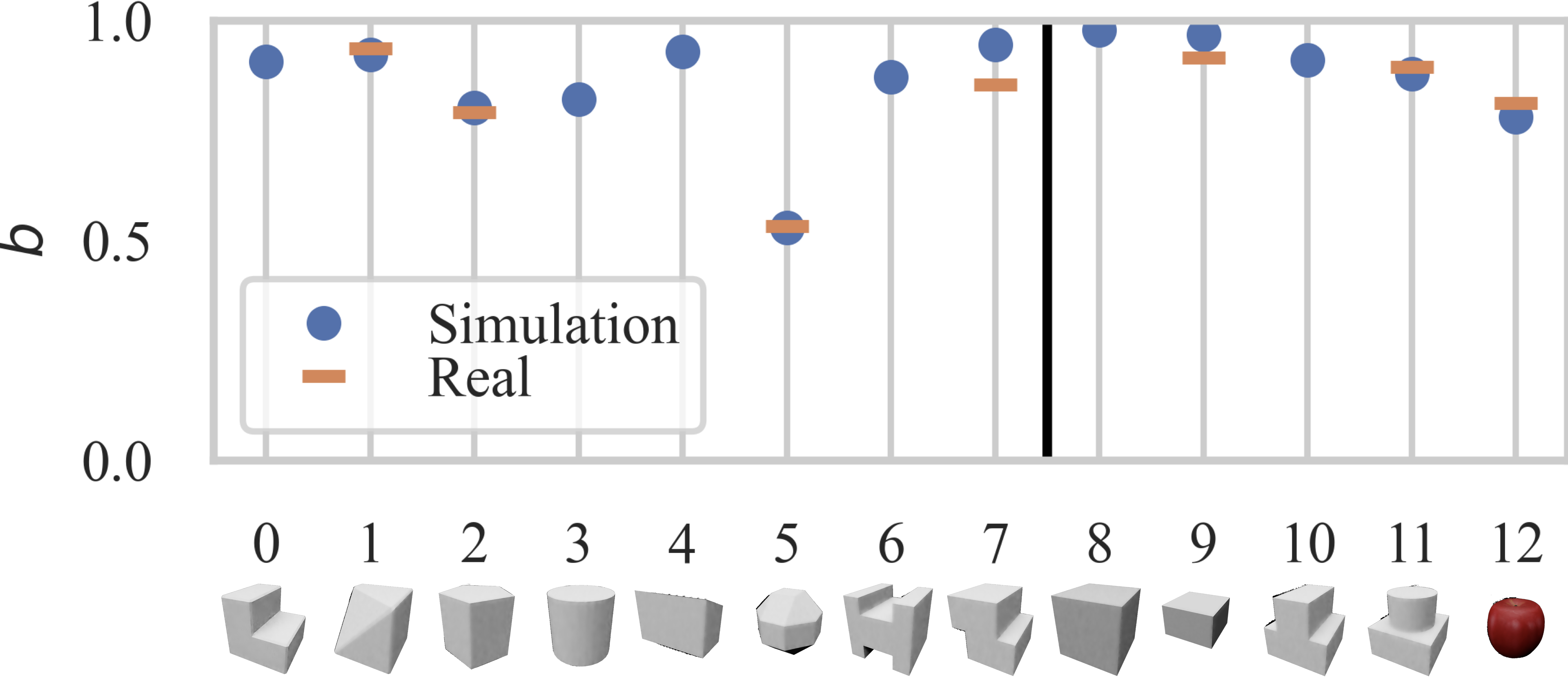}
    \caption{
        Real-world and simulation success rates $b$ for individual objects (nominal object scaling).
        Experiments are run with the same policy and estimator with \textit{VecBPS} shape representation.
        In simulation, we plot the mean over 256 environments running all 24 goals. 
        For the real-world experiments, we picked two easy and two hard training objects plus three \textsc{OOD} objects. 
        We evaluated each goal twice.
    }
    \label{fig:success_plot}    
\end{figure}

In \cref{fig:success_plot}, we benchmark the success rates of the tactile agent for the \textsc{Geometric 8} training objects at aspect ratio 1 and the \textsc{OOD} objects.
We see no clear performance difference between the training objects and the \textsc{OOD} object set, indicating that the policy and estimator can generalize well to novel objects with similar properties. 

The success rates for objects 1, 8, 9, and 12 (L, cube, cuboid, apple) are similar to the state-of-the-art results for the respective objects reported in~\citet{Rostel2023-nc}, where we trained a separate agent for each object.
Interestingly, our agent can successfully adapt its control strategy when manipulating object 12, the \textsc{OOD} apple, which requires an especially deliberate manipulation strategy that avoids estimator drift~\cite{Rostel2023-nc}.
We encourage the reader to inspect the supplementary video and the project website to see the qualitatively different manipulation patterns produced by the same agent conditioned on various shapes.

The success rate for the apple is even significantly higher than for object 5, the 26-sided ``sphere''.
We assume the 45-degree angles are somewhat adversarial for the tactile state estimator because it is difficult to guess which side a finger touches on the object. 
However, unlike for round objects, it matters a lot since the object may snap to one side or the other, depending on the exact contact location.
The second mode in object 5's violin (\cref{fig:violinplot}) is likely caused by this snapping effect. 
The estimator might then realize that the finger is placed properly on a face but cannot tell which one.

\subsection{Real-World Experiments}

To validate our simulation results, we conduct real-world experiments with the torque-controlled DLR-Hand II~\cite{Butterfass2001}, with calibrated kinematics~\cite{Tenhumberg2023-Hand}.
The real-world experimental protocol follows a similar procedure as in simulation:
The operator first passes in the object in its canonical orientation.
Then, the hand is closed around the object by commanding a fixed set of ``grasping'' target angles for the impedance controller.
Finally, we run the agent for 10\,s and evaluate if the object was correctly reoriented. 
Due to the 90-degree discretization of the benchmark task, it is usually easy to judge if the final angle error is below the 0.4\,rad threshold even without a visual tracking system.

Out of the large number of possible objects to be tested, we choose to evaluate the two \textsc{Geometric 8} objects with the highest and two with the lowest success rates in simulation, as well as three of the \textsc{OOD} objects.
For each object, we evaluate all of the 24 benchmark goals twice and report the aggregated success rates.
The results shown in \cref{fig:success_plot} indicate that real-world performance closely matches the simulation results.

The supplementary video and the project website show the real-world experiments.

\subsection{Assessing the Limits of \textsc{OOD} Generalization}

\begin{figure}
    \centering
    \vspace{3mm}
    \includegraphics[width=3.4in]{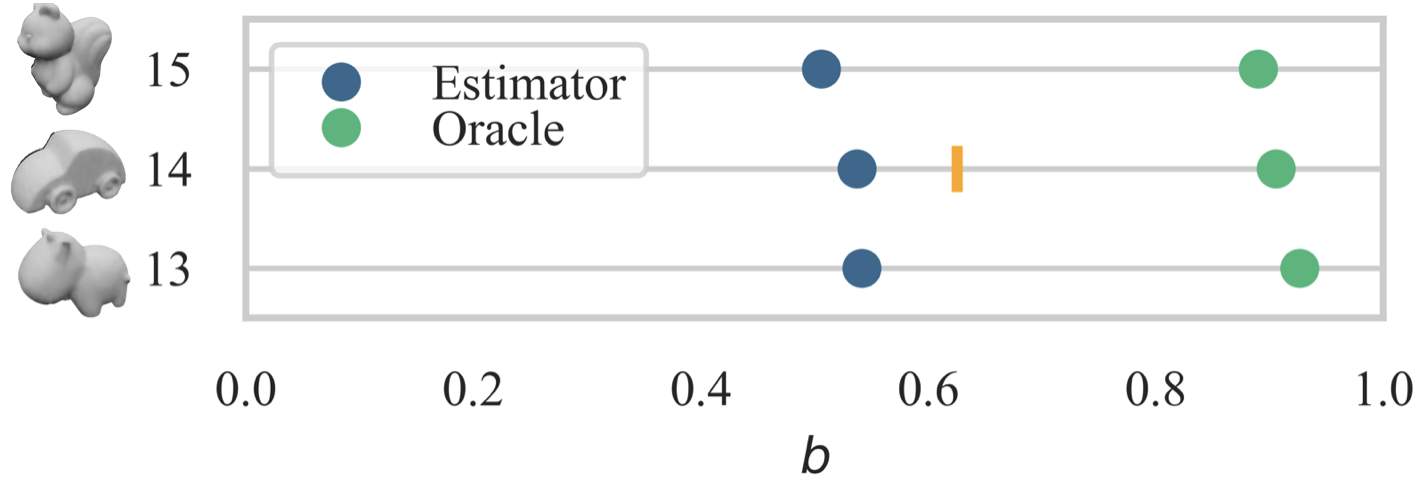}
    \caption{
        Success rates for additional complex, \textsc{OOD} objects. 
        We use the same experiment settings as in \cref{fig:success_plot} and additionally plot the oracle policy performance for reference.
    }
    \label{fig:miscnet_plot}    
\end{figure}

In \cref{fig:miscnet_plot} we test the limits of our agents, trained on randomized geometric objects, by evaluating it, in simulation, on three complex objects (13-15 originally introduced in the context of vision-based in-hand manipulation in~\citet{Chen2023-mq}).
We also check the real-world performance of object 14, which again matches well with the simulation.
While the oracle policy handles these objects surprisingly well, a performance gap is seen for the purely tactile agent.
Our hypothesis for this observation is that, due to the complex nature of these objects (small features, high curvature), the sensitivity of the control strategy to even small inaccuracies in the estimated object pose is even higher. 
Uncontrolled object movements resulting from this can lead to permanent loss of observability, as the object pose can not be determined unambiguously by a tactile estimator~\cite{Rostel2023-nc}.
We expect that the ability to sense small features as provided by tactile skin, could be necessary to reorient these objects with greater robustness.

\subsection{Sensitivity to Shape Conditioning}

\begin{figure}
    \centering
    \vspace{3mm}
    \includegraphics{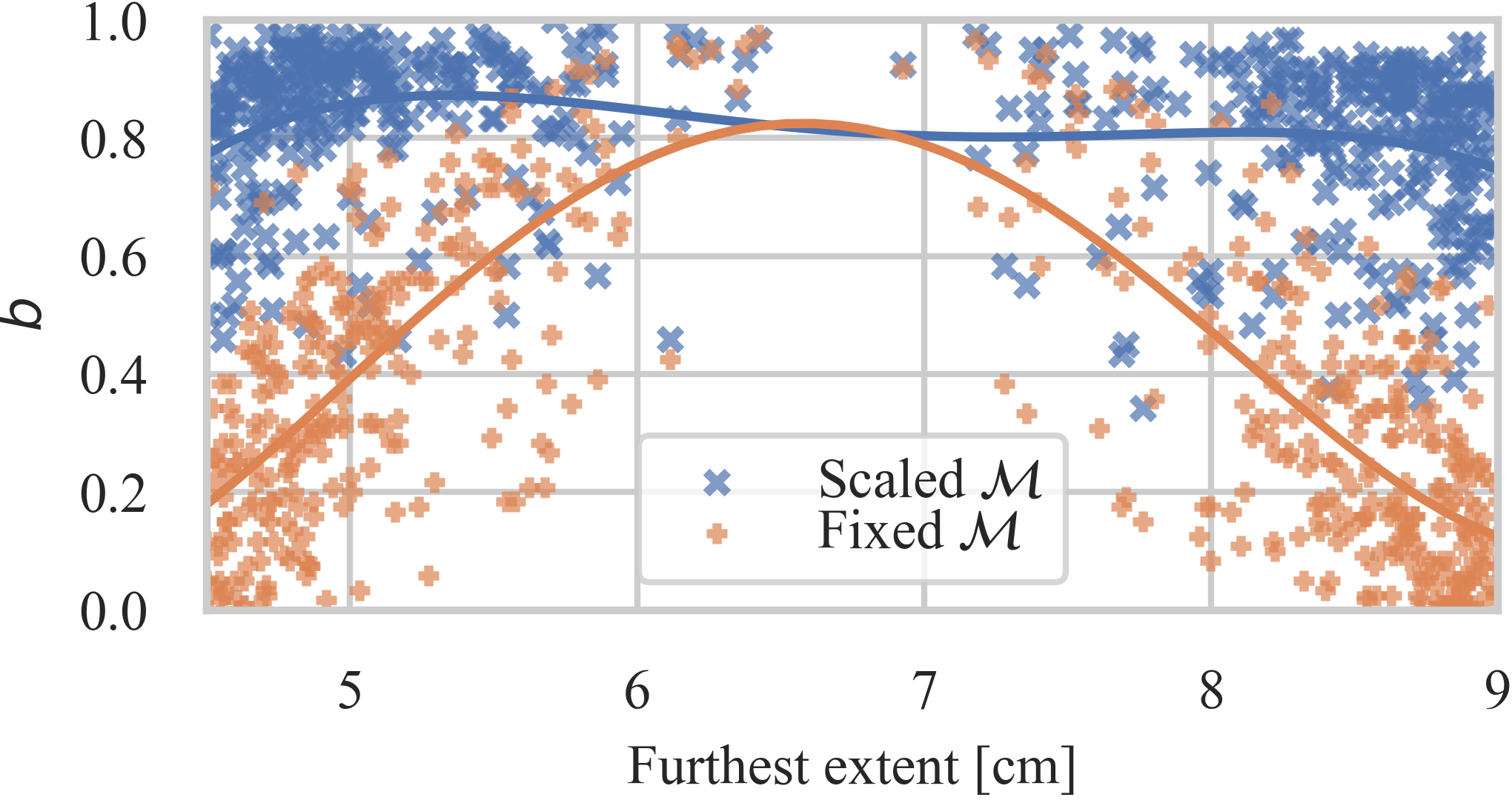}
    \caption{
        Success rates $b$ for randomly scaled training objects. The x-axis indicates the extent which is furthest away from the median (6.75\,cm) for each object.
        Each sample is the mean of one environment over five evaluations for all 24 goals (64 for each training object, totaling 512 samples).
        The lines are polynomials of degree 4 fitted to the samples.
        We compare the same \textit{VecBPS} agent (with estimator) when passing in the correctly scaled $\mathcal{M}$ versus always passing in a fixed $\mathcal{M}$ with median extents.
        Note that when the furthest extent is far away from the median, the agent's success rate conditioned on the fixed~$\mathcal{M}$ decreases quickly, while conditioned on the correctly scaled~$\mathcal{M}$, the agent copes well with larger aspect ratios.
    }
    \label{fig:extent_scatterplot}
\end{figure}

\begin{figure}
    \centering
    \vspace{3mm}
    \includegraphics{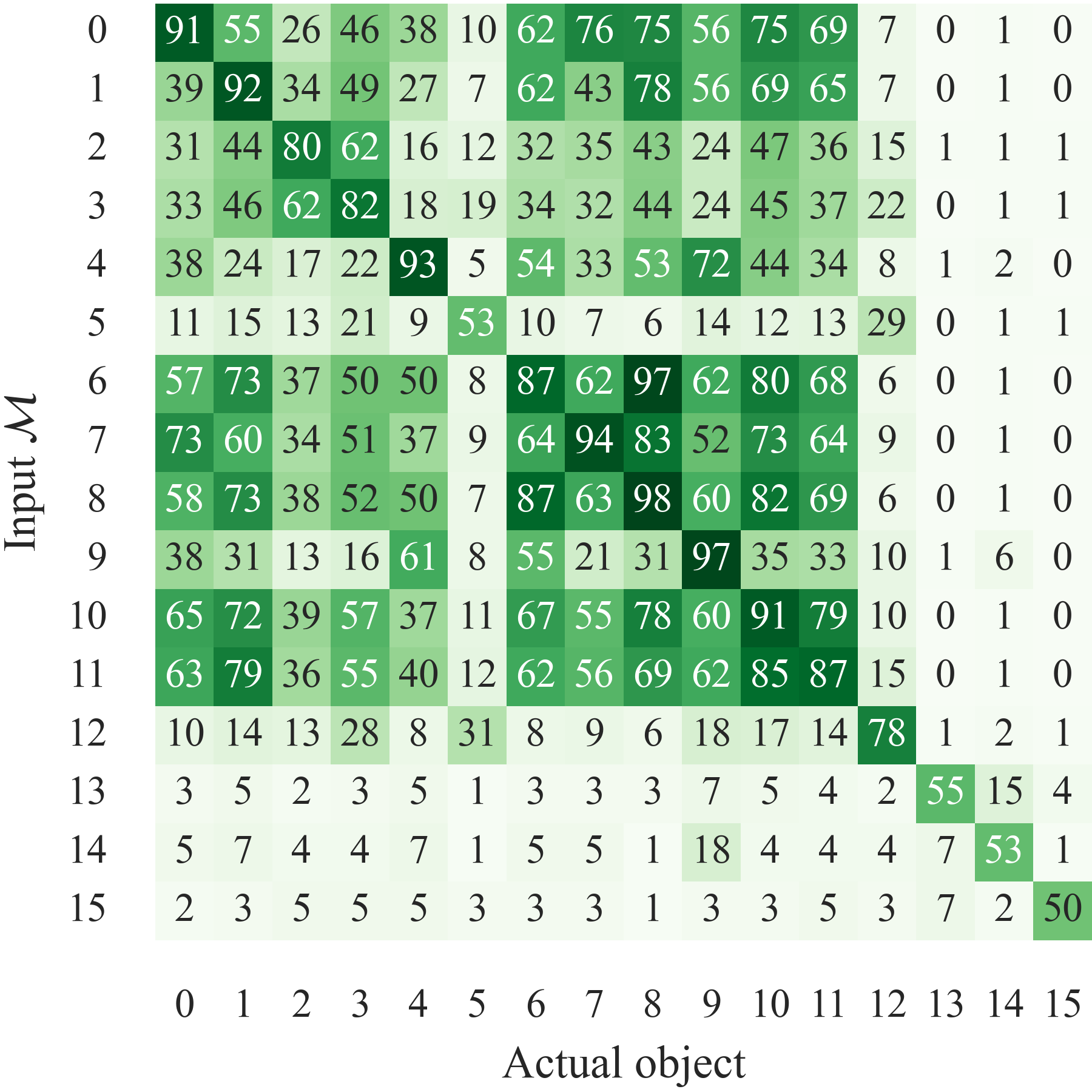}
    \caption{
        ``Confusion matrix'' shows the success rates in \% for individual objects (nominal object scaling).
        Each entry is the mean over 256 environments running all 24 goals.
        The row index indicates which mesh $\mathcal{M}$ the policy and estimator were conditioned on, and the column index indicates which object was actually manipulated.
    }
    \label{fig:matrix_plot}
\end{figure}

We test the agent's performance when provided with an incorrect mesh to further understand the models' ability to adapt based on shape conditioning.
For this, we conduct two experiments, testing the influence of the object scale and geometry separately.
In \cref{fig:extent_scatterplot}, we evaluate the success rate when provided with a mesh that does not account for the actual object extents. 
We fix the object scaling for the experiment in \cref{fig:matrix_plot} but provide the agent with an entirely different mesh.
In both cases, we observe a significant drop in performance when the shape information is incorrect, confirming that manipulation strategies indeed heavily depend on the shape conditioning.

It is interesting to see how some objects correlate quite strongly, such as objects 2 and 3 (cylinder and pentagon), objects 11 and 12 (T and emergency switch), or objects 6 and 8 (H and cube). 
Still, especially for the out-of-distribution objects 12-15, conditioning the agent on the correct shape is vital.

Note that all objects perform best when conditioned on the correct shape (the diagonal). 
Only object 6 works just as well when conditioned on the cube mesh (object 8), which might indicate that the spatial resolution of our basis point set is not high enough to resolve smaller concave features.

\subsection{Ablations}
To understand the effect of different design decisions on the final performance when training with the estimator, we perform a series of ablation studies. 
\cref{tab:estim_ablations} shows the success rates and final angle to goal relative to the performance of the base model (with \textit{VecBPS} encoding) evaluated in \cref{sec:experiments_estim}. 
Not giving the policy the estimated uncertainty $\sigma$ or stopping the gradient through the BPS calculation when training the state estimator leads to slight but notable decreases in performance.
Omitting shape information from the system leads to a drop in performance even larger than observed for the \textit{oracle} case (\cref{sec:experiments_oracle}), indicating that shape information is crucial for learning accurate state estimation.
Interestingly, using the BPS distances only does not significantly affect the final performance.

\begin{table}
    \centering
    \caption{Training with Estimator - Ablations}
    \begin{tabular}{l c c }
        \toprule
        \quad & \makecell{relative\\ success rate} &\makecell{relative\\ final angle to goal} \\
        \midrule
        No $\sigma$ input to $\pi$ & $-2.6$\%  & $+12$\%\\
        Stop gradient through $\mathcal{B}$ & $-1.7$\% & $+5$\%\\
        No shape info &  $-30$\%& $+89$\% \\
        \textit{BPS} (distance only) & $-0$\%& $-0.2$\%\\
        \bottomrule
    \end{tabular}
    \label{tab:estim_ablations}
\end{table}

\section{Conclusion and Outlook}

We presented a method for learning a shape-conditioned policy and state estimator for in-hand manipulation.
Using our agent, we were the first to autonomously reorient various objects to specified target orientations, purely based on torque and position feedback, all using a single agent.
To this end, we compared different shape representations and found that basis point sets computed for the transformed object meshes are a well-suited representation for in-hand manipulation.
Although we could show high success rates for many objects, both seen and unseen during training, we also observed limitations, especially regarding objects with small features. 
In future work, we plan to address these limitations by incorporating tactile skin as an additional sensor modality, with which we hope to scale to more objects and tasks.

\footnotesize
\bibliographystyle{IEEEtranN-modified}
\bibliography{IEEEabrv, bibliography}


\end{document}